\documentclass[11pt,a4paper]{article}
\usepackage{acl2016}
\usepackage{times}
\usepackage[utf8]{inputenc}
\usepackage{url}
\usepackage{latexsym}
\usepackage{amsmath}
\usepackage{amsfonts}
\usepackage{amssymb}
\usepackage{graphicx}
\usepackage{booktabs}
\usepackage{multirow}
\usepackage{float}
\usepackage{enumitem}
\usepackage{pbox}
\usepackage{CJKutf8}
\usepackage[font=footnotesize, labelfont=bf]{caption}
\aclfinalcopy


\makeatletter
\newcommand\thefontsize[1]{{#1 The current font size is: \f@size pt\par}}
\makeatother

\makeatletter
\def\citealt{\def\citename##1{{\frenchspacing##1}, }\@internalcitec}

\def\@citexc[#1]#2{\if@filesw\immediate\write\@auxout{\string\citation{#2}}\fi
  \def\@citea{}\@citealt{\@for\@citeb:=#2\do
    {\@citea\def\@citea{;\penalty\@m\ }\@ifundefined
       {b@\@citeb}{{\bf ?}\@warning
       {Citation `\@citeb' on page \thepage \space undefined}}%
{\csname b@\@citeb\endcsname}}}{#1}}

\def\@internalcitec{\@ifnextchar [{\@tempswatrue\@citexc}{\@tempswafalse\@citexc[]}}

\def\@citealt#1#2{{#1\if@tempswa, #2\fi}}
\makeatother
\newcommand{\mb}{\mathbf}
\newcommand{\R}{\mathbb{R}}
\newcommand{\pmi}{\textrm{PPMI}}
\newcommand{\ind}{\mathbb{I}}
\newcommand{\engall}{\textsc{EngAll} }
\newcommand{\engfic}{\textsc{EngFic} }
\newcommand{\coha}{\textsc{COHA} }

\author{William L.\@ Hamilton, Jure Leskovec, Dan Jurafsky \\
Department of Computer Science, Stanford University, Stanford CA, 94305\\
\texttt{wleif,jure,jurafsky@stanford.edu}}
\title{Diachronic Word Embeddings Reveal Statistical Laws of\\ Semantic Change}
\begin{document}
\maketitle
\begin{abstract}

%\begin{small}

%\renewcommand{\baselinestretch}{1.5}

%\setlength{\baselineskip}{11.5pt}
Understanding how words change their meanings over time is key to models of
language and cultural evolution,  but historical data on meaning is scarce,
making theories hard to develop and test.
Word embeddings show promise as a diachronic tool,
but have not been carefully evaluated.
We develop a robust methodology for quantifying semantic change by
evaluating word embeddings (PPMI, SVD, word2vec) against known historical changes.
We then use this methodology to reveal statistical laws of semantic evolution. Using six
historical corpora spanning four languages and two centuries, we
propose two quantitative laws of semantic change: (i) the {\em law of
conformity}---the rate of semantic change scales with an inverse power-law
of word frequency; (ii) the {\em law of innovation}---independent of frequency,
words that are more polysemous have higher rates of semantic change.

%\end{small}
%has made it hard to develop or test theories of semantic change.  
%of language and cultural evolution,  but we lack good historical data on meaning, making
\end{abstract}

\section{Introduction}

Shifts in word meaning exhibit systematic regularities \cite{breal_essai_1897,ullmann_semantics:_1962}.
The rate of semantic change, for example, is higher in 
some words than others \cite{blank_why_1999} --- compare
the stable semantic history of \textit{cat} (from Proto-Germanic \textit{kattuz}, ``cat'') to
the varied meanings of English \textit{cast}: ``to mould'', ``a collection of actors', ``a hardened bandage'', etc.\@ (all from Old Norse \textit{kasta}, ``to throw'', \citealt{simpson_oxford_1989}).

Various hypotheses have been offered about such regularities in semantic change,
such as an increasing subjectification of meaning, or the grammaticalization
of inferences (e.g., \citealt{geeraerts_diachronic_1997,blank_why_1999,traugott_regularity_2001}).
%,hopper_grammaticalization_2003}),

But many core questions about semantic change remain unanswered.
One is the role of {\em frequency}. Frequency plays a key role in other linguistic changes, 
associated sometimes with faster change---sound changes like lenition occur in more frequent words---and sometimes with slower change---high frequency words are more resistant to morphological regularization
\cite{bybee_frequency_2007,pagel_frequency_2007,lieberman_quantifying_2007}. 
What is the role of word frequency in meaning change?

Another unanswered question is the relationship between semantic change and {\em polysemy}.
Words gain senses over time as they semantically drift \cite{breal_essai_1897,wilkins_part_1993,hopper_grammaticalization_2003},
and polysemous words\footnote{We use `polysemy' here to refer 
to related senses as well as rarer cases of accidental homonymy.}
occur in more diverse contexts, affecting lexical access speed \cite{adelman} and
rates of L2 learning \cite{crossley2010}.
But we don't know whether the diverse contextual use of polysemous words makes them  more or less likely to undergo change
\cite{geeraerts_diachronic_1997,winter_cognitive_2014,xu_historical_2015}.
Furthermore, polysemy is strongly correlated with frequency---high frequency words have more senses \cite{zipf_meaning-frequency_1945,ilgen_investigation_2007}---so understanding how polysemy relates to semantic change requires controling for word frequency.  

%Word senses have a Zipfian distribution, meaning that most senses of words are quite rare \cite{kilgariff},
%
%And it has been argued that ``words become semantically extended by being used in diverse contexts'' \cite{winter_cognitive_2014}, which implies that highly polysemous words are more likely to undergo semantic change 
%Both these notions suggest that polysemy should be positively correlated with rates of semantic change. 
%In sum, the dominant hypothesis in the literature is that polysemy is positively correlated with semantic change. 

\begin{figure*}[t]
\centering
\includegraphics[width=\textwidth]{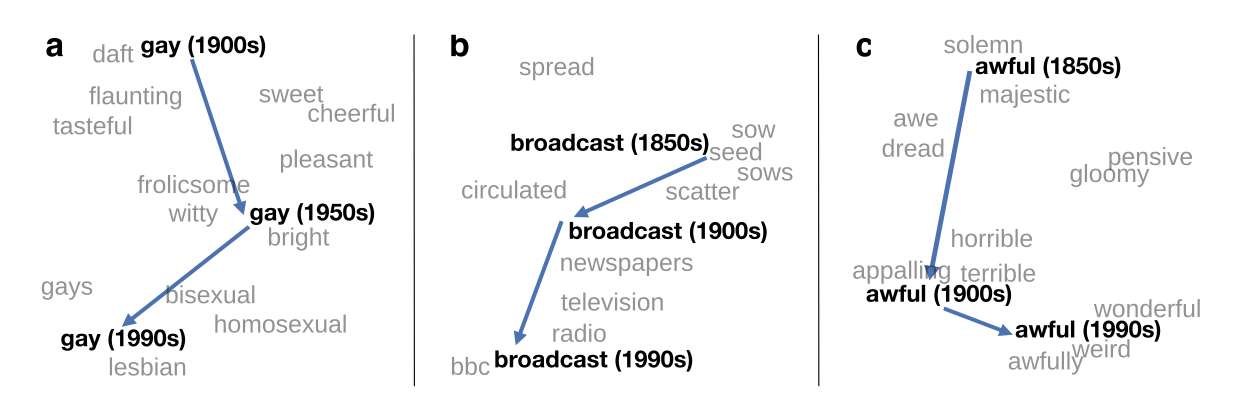}
\vspace*{-10pt}
\caption{Two-dimensional visualization of semantic change in English using SGNS vectors.\footnotemark    
$\:$\textbf{a}, The word \textit{gay} shifted from meaning ``cheerful'' or ``frolicsome'' to referring to homosexuality. \textbf{b}, In the early 20th century \textit{broadcast} referred to ``casting out seeds''; with the rise of television and radio its meaning shifted to ``transmitting signals''. \textbf{c}, \textit{Awful} underwent a process of pejoration, as it shifted from meaning ``full of awe'' to meaning ``terrible or appalling'' \protect\cite{simpson_oxford_1989}.} 
\label{wordpaths}
\vspace{-10pt}
\end{figure*}

Answering these questions requires new methods that can go beyond the case-studies of a few words
(often followed over widely different time-periods) that are our most common diachronic data
\cite{breal_essai_1897,ullmann_semantics:_1962,blank_why_1999,hopper_grammaticalization_2003,traugott_regularity_2001}.
One promising avenue is the use of distributional semantics, in which
words are embedded in vector spaces according to their co-occurrence relationships
\cite{bullinaria_extracting_2007,turney_frequency_2010}, and 
the embeddings of words are then compared across time-periods. 
This new direction has been effectively demonstrated in a number of case-studies 
\cite{sagi_tracing_2011,wijaya_understanding_2011,gulordava_distributional_2011,jatowt_framework_2014}
and used to perform large-scale linguistic change-point detection \cite{kulkarni_statistically_2014}
as well as to test a few specific hypotheses,
such as whether English synonyms tend to change meaning in similar ways \cite{xu_computational_2015}.
However, these works employ widely different embedding approaches
%don't benchmark on carefully curated corpora like COHA \cite{coha},
and test their approaches only on English.

%Our current understanding of these regularities in semantic change, however, is based only on small amounts of data:
%individual case-studies of a few words, selected by hand and often followed over widely different periods in time 
%\cite{breal_essai_1897,ullmann_semantics:_1962,blank_why_1999,hopper_grammaticalization_2003,traugott_regularity_2001}.
%This traditional methodology alone is insufficient for quantifying the precise nature of the relationship between frequency, polysemy, and semantic change.
%
%For example, it is known that polysemy is positively correlated with frequency --- i.e., higher frequency words are more polysemous \cite{zipf_meaning-frequency_1945,ilgen_investigation_2007}; thus any analysis relating polysemy to semantic change will need to statistically control for the effect of frequency.  

In this work, we develop a robust methodology for quantifying semantic change using embeddings by comparing state-of-the-art approaches (PPMI, SVD, word2vec) on novel benchmarks.

We then apply this methodology in a large-scale cross-linguistic analysis using 6 corpora spanning 200 years and 4 languages
(English, German, French, and Chinese).
Based on this analysis, we propose two statistical laws relating frequency and polysemy to semantic change:
\begin{itemize}[itemsep=0pt,topsep=0pt, parsep=0pt]
	\item
		\textbf{The law of conformity:} Rates of semantic change scale with a negative power of word frequency. 
	\item
		\textbf{The law of innovation:} After controlling for frequency, polysemous words have significantly higher rates of semantic change. 
	\end{itemize}
\footnotetext{Appendix B details the visualization method.}
\section{Diachronic embedding methods}\label{methods}
\begin{table*}[t!]
\centering
{
\begin{tabular}{lllrr} 
\toprule  
Name & Language & Description & Tokens & Years \\ 
\midrule 
\engall & English & Google books (all genres) & $8.5 \times 10^{11}$ & 1800-1999 \\
\engfic & English & Fiction from Google books & $7.5 \times 10^{10}$ & 1800-1999 \\
\coha & English & Genre-balanced sample & $4.1 \times 10^{8}$ & 1810-2009 \\
\textsc{FreAll} & French &  Google books (all genres) & $1.9 \times 10^{11}$& 1800-1999 \\
\textsc{GerAll} & German &  Google books (all genres) & $4.3 \times 10^{10}$& 1800-1999 \\
\textsc{ChiAll} & Chinese &  Google books (all genres) & $6.0 \times 10^{10}$& 1950-1999 \\
\bottomrule
\end{tabular} 
}
\caption{Six large historical datasets from various languages and sources are used. }
\label{datasets-table}
\end{table*}

The following sections outline how we construct diachronic (historical) word embeddings,
by first constructing embeddings in each time-period and then aligning them over time,
and the metrics that we use to quantify semantic change.
All of the learned embeddings and the code we used to analyze them are made publicly available.\footnote{{\scriptsize\url{http://nlp.stanford.edu/projects/histwords}}}

\subsection{Embedding algorithms}
We use three methods to construct word embeddings within each time-period: PPMI, SVD, and SGNS (i.e., word2vec).\footnote{Synchronic applications of these three methods are reviewed in detail in \newcite{levy_improving_2015}.}
These distributional methods represent each word $w_i$ by a vector $\mb{w}_i$ that captures information 
about its co-occurrence statistics. 
These methods operationalize the `distributional hypothesis' that word semantics are implicit in co-occurrence relationships \cite{Harris54,firth_synopsis_1957}. 
The semantic similarity/distance between two words is approximated by the cosine similarity/distance between their vectors \cite{turney_frequency_2010}.

\subsubsection{PPMI}

In the PPMI representations, the vector embedding for word $w_i \in \mathcal{V}$ contains the positive point-wise mutual information (PPMI) values between $w_i$ and a large set of pre-specified `context' words. 
The word vectors correspond to the rows of the matrix $\mb{M}^{\textrm{PPMI}} \in \R^{|\mathcal{V}| \times |\mathcal{V}_C|}$ with entries given by
\begin{equation}
\mb{M}^{\textrm{PPMI}}_{i,j} = \max\left\lbrace\log\left(\frac{\hat{p}(w_i,c_j)}{\hat{p}(w)\hat{p}(c_j)}\right)-\alpha ,0\right\rbrace,
\end{equation}
where $c_j \in \mathcal{V}_C$ is a context word and $\alpha > 0$ is a negative prior, which provides a smoothing bias \cite{levy_improving_2015}. 
The $\hat{p}$ correspond to the smoothed empirical probabilities of word \mbox{(co-)occurrences} within fixed-size sliding windows of text. 
Clipping the PPMI values above zero ensures they remain finite and has been shown to dramatically improve results \cite{bullinaria_extracting_2007,levy_improving_2015}; intuitively, this clipping ensures that the representations emphasize positive word-word correlations over negative ones. 

\subsubsection{SVD}

SVD embeddings correspond to low-dimensional approximations of the PPMI embeddings learned via singular value decomposition \cite{levy_improving_2015}. 
The vector embedding for word $w_i$ is given by 
\begin{equation}
\mb{w}^{\textrm{SVD}}_{i} = \left(\mb{U}\mb{\Sigma}^\gamma\right)_{i},
\end{equation}
 where $\mb{M}^{\textrm{PPMI}} = \mb{U}\mb{\Sigma}\mb{V}^\top$ is the truncated singular value decomposition of $\mb{M}^{\textrm{PPMI}}$ and $\gamma \in [0,1]$ is an eigenvalue weighting parameter. Setting $\gamma <1$ has been shown to dramatically improve embedding qualities \cite{turney_frequency_2010,bullinaria_extracting_2012}. 
 This SVD approach can be viewed as a generalization of Latent Semantic Analysis \cite{landauer_solution_1997}, where the term-document matrix is replaced with $\mb{M}^{\textrm{PPMI}}$. 
 Compared to PPMI, SVD representations can be more robust, as the dimensionality reduction acts as a form of regularization. 
 
\subsubsection{Skip-gram with negative sampling}

SGNS `word2vec' embeddings are optimized to predict co-occurrence relationships using an approximate objective known as `skip-gram with negative sampling' \cite{mikolov_distributed_2013}. In SGNS, each word $w_i$ is represented by two dense, low-dimensional vectors: a word vector ($\mb{w}^{\textrm{SGNS}}_i$) and context vector ($\mb{c}^{\textrm{SGNS}}_i)$. These embeddings are optimized  via stochastic gradient descent so that 
\begin{equation}\label{sgns}
 \hat{p}(c_i | w_i) \propto \exp(\mb{w}^{\textrm{SGNS}}_i\cdot\mb{c}^{\textrm{SGNS}}_j),
\end{equation}
where $p(c_i | w_i)$ is the empirical probability of seeing context word $c_i$ within a fixed-length window of text, given that this window contains $w_i$.
The SGNS optimization avoids computing the normalizing constant in \eqref{sgns} by randomly drawing `negative' context words, $c_n$, for each target word and ensuring that $\exp(\mb{w}^{\textrm{SGNS}}_i\cdot\mb{c}^{\textrm{SGNS}}_n)$ is small for these examples. 

SGNS has the benefit of allowing incremental initialization during learning, where the embeddings for time $t$ are initialized with the embeddings from time $t-\Delta$ \cite{kim_temporal_2014}.

%\subsubsection{Hyperparameter settings}

 %Appendix A details the full set of hyperparameter settings and implementations employed. 

\subsection{Datasets, pre-processing, and hyperparameters}
We trained models on the 6 datasets described in Table \ref{datasets-table}, taken from Google N-Grams \cite{lin_syntactic_2012} and the \coha  corpus \cite{davies_corpus_2010}.
The Google N-Gram datasets are extremely large (comprising ${\approx}6\%$ of all books ever published), but they also contain many corpus artifacts due, e.g., to shifting sampling biases over time \cite{pechenick_characterizing_2015}. 
In contrast, the \coha corpus was carefully selected to be genre-balanced and representative of American English over the last 200 years, 
though as a result it is two orders of magnitude smaller.
The \coha corpus also contains pre-extracted word lemmas, which we used to validate that our results hold at both the lemma and raw token levels. 
 All the datasets were aggregated to the granularity of decades.\footnote{The 2000s decade of the Google data was discarded due to shifts in the sampling methodology \cite{michel_quantitative_2011}.}
 %\footnote{We kept the full context set in the PPMI vectors however, as it was found to improve results for that method.}  
 %Some aspects of our analysis rely on detailed part-of-speech (POS) tags (e.g., indicating proper nouns), which are not present in the coarse-grained Google POS data. We thus extracted majority-vote POS tags from the third-party sources indicated in Table \ref{datasets-table}.  

We follow the recommendations of \newcite{levy_improving_2015} in setting the hyperparameters for the embedding methods, though preliminary experiments were used to tune key settings.
For all methods, we used symmetric context windows of size 4 (on each side).
For SGNS and SVD, we use embeddings of size 300. See Appendix A for further implementation and pre-processing details.

\subsection{Aligning historical embeddings}

In order to compare word vectors from different time-periods we must ensure that the vectors are aligned to the same coordinate axes. 
Explicit PPMI vectors are naturally aligned, as each column simply corresponds to a context word. 
Low-dimensional embeddings will not be naturally aligned due to the non-unique nature of the SVD and the stochastic nature of SGNS.
In particular, both these methods may result in arbitrary orthogonal transformations, which do not affect pairwise cosine-similarities within-years but will preclude comparison of the same word across time.
Previous work circumvented this problem by either avoiding low-dimensional embeddings (e.g., \citealt{gulordava_distributional_2011,jatowt_framework_2014}) or by performing heuristic local alignments per word \cite{kulkarni_statistically_2014}.
%The procedure below provides an efficient and principled way to globally align the vectors.\footnote{In principle, one could use our approach to perform local alignments per word, similar to the heuristic method of \newcite{kulkarni_statistically_2014}, but preliminary experiments revealed no clear benefits to this approach.}  
	
We use orthogonal Procrustes to align the learned  low-dimensional embeddings.
Defining $\mb{W}^{(t)} \in \R^{d \times |\mathcal{V}|}$ as the matrix of word embeddings learned at year $t$, we align across time-periods while preserving cosine similarities by optimizing:
\begin{equation}\label{procrustes}
\mb{R}^{(t)} = \arg\min_{\mb{Q}^\top\mb{Q} = \mb{I}} \| \mb{Q}\mb{W}^{(t)} - \mb{W}^{(t+1)} \|_F,
\end{equation}
with $\mb{R}^{(t)} \in \R^{d \times d}$. The solution corresponds to the best rotational alignment and can be obtained efficiently using an application of SVD \cite{schonemann_generalized_1966}.
	%Again, we perform the alignment only over words that occurred more than certain threshold (500 for the Google data, 100 for COHA).

\subsection{Time-series from historical embeddings}

Diachronic word embeddings can be used in two ways to quantify semantic change: (i) we can measure changes in pair-wise word similarities over time, or (ii) we can measure how an individual word's embedding shifts over time. 

\paragraph{Pair-wise similarity time-series} 
	Measuring how the cosine-similarity between pairs of words changes over time allows us to test hypotheses about specific linguistic or cultural shifts in a controlled manner. 
	We quantify shifts by computing the similarity time-series 
\begin{equation}
	s^{(t)}(w_i,w_j) = \textrm{cos-sim}(\mb{w}^{(t)}_i,\mb{w}^{(t)}_j)
\end{equation}	
between two words $w_i$ and $w_j$ over a time-period $(t,...,t+\Delta$).
We then measure the Spearman correlation ($\rho$)  of this series against time, which allows us to assess the magnitude and significance of pairwise similarity shifts; since the Spearman correlation is non-parametric, this measure essentially detects whether the similarity series increased/decreased over time in a significant manner, regardless of the `shape' of this curve.\footnote{Other metrics or change-point detection approaches, e.g. mean shifts \cite{kulkarni_statistically_2014} could also be used.}
%Figure \ref{wordpaths} shows 2D-visualizations of some well-known shifts in English\footnote{Appendix B details the visualization method.} %, constructed using pairwise similarities and t-SNE \cite{van_der_maaten_visualizing_2008}.	
	%	For example, using pairwise similarities from SVD vectors we can characterize well-known shifts in different languages:
	%	in German, \textit{Pappe} (formerly ``paste'') shifted significantly\footnote{All reported statistics are significant at the $p<0.05$ level unless indicated otherwise.} closer to \textit{Papier} (``paper''), as it acquired its modern meaning of ``cardboard'' over the past two centuries  ($\rho=0.90$);
	%		in French, \textit{fichier} (``file'') shifted closer to \textit{informatique} (``data processing'') over the latter half of the twentieth century ($\rho=1.0$);
	%and in the last 50 years of Chinese text, we see that\begin{CJK*}{UTF8}{gbsn} 病毒 \end{CJK*}  (``virus'') became closer to\begin{CJK*}{UTF8}{gbsn} 电脑 \end{CJK*} (``computer'', $\rho=0.89$).	F	

\paragraph{Measuring semantic displacement}\label{displacement}

After aligning the embeddings for individual time-periods, we can use the aligned word vectors to compute the semantic displacement that a word has undergone during a certain time-period. 
In particular, we can directly compute the cosine-distance between a word's representation for different time-periods, i.e.\@ $\textrm{cos-dist}(\mb{w}_t, \mb{w}_{t+\Delta})$, as a measure of semantic change. 
We can also use this measure to quantify `rates' of semantic change for different words by looking at the displacement between consecutive time-points.

\section{Comparison of different approaches}

We compare the different distributional approaches on a set of
benchmarks designed to test their scientific utility.  We evaluate
both their {\em synchronic} accuracy (i.e., ability to capture word similarity within
individual time-periods) and their {\em diachronic} validity 
(i.e., ability to quantify semantic changes over time).

\subsection{Synchronic Accuracy}

We evaluated the synchronic (within-time-period) accuracy of the methods using a standard modern benchmark and the 1990s portion of the \engall data. 
On \newcite{bruni_distributional_2012}'s MEN similarity task of matching human judgments of word similarities, SVD performed best ($\rho=0.739$), followed by PPMI ($\rho=0.687$) and SGNS ($\rho=0.649$).
These results echo the findings of \newcite{levy_improving_2015}, who found SVD to perform best on similarity tasks while SGNS performed best on analogy tasks (which are not the focus of this work). 
%Of course, it is difficult to draw firm conclusions given the dependency of hyperparameters and the dataset etc., but it appears that SGNS does not as fair well with this synchronic task.
%	Qualitatively, however, we found the SGNS nearest neighbors to be very high quality, which is reflected in its performance in some of the historical tasks below. 
%	It appears that PPMI, and by extension SVD, benefits from its sensitivity to rare co-occurrences in the benchmark task but that this also leads to spurious and uninformative nearest neighbors. 

\subsection{Diachronic Validity}

We evaluate the diachronic validity of the methods on two historical semantic tasks: detecting known shifts and discovering shifts from data.
For both these tasks, we performed detailed evaluations on a small set of examples (28 known shifts and the top-10 ``discovered'' shifts by each method).
Using these reasonably-sized evaluation sets allowed the authors to evaluate each case rigorously using existing literature and historical corpora.
%\footnote{Crowdsourced evaluation is not applicable in this setting due to the subtleties involved in judging semantic changes.} 

\paragraph{Detecting known shifts.}
\begin{table*}
\centering
{\small
\begin{tabular}{llllr} 
\toprule  
Word & Moving towards & Moving away & Shift start & Source \\ 
\midrule 
gay & homosexual, lesbian & happy, showy &  ca 1950 & \cite{kulkarni_statistically_2014}\\
fatal &illness, lethal & fate, inevitable &  $<$1800 & \cite{jatowt_framework_2014}\\
awful & disgusting, mess & impressive, majestic & $<$1800 & \cite{simpson_oxford_1989}\\
nice & pleasant, lovely & refined, dainty &  ca 1890 & \cite{wijaya_understanding_2011}\\
broadcast & transmit, radio & scatter, seed &  ca 1920 & \cite{jeffers_principles_1979}\\
monitor & display, screen & --- &  ca 1930 & \cite{simpson_oxford_1989}\\
record & tape, album & --- &  ca 1920 & \cite{kulkarni_statistically_2014}\\
guy & fellow, man & --- &  ca 1850 & \cite{wijaya_understanding_2011}\\
call & phone, message & --- & ca 1890 & \cite{simpson_oxford_1989}\\
\bottomrule
\end{tabular} 
}
\caption{Set of attested historical shifts used to evaluate the methods. The examples are taken from previous works on semantic change and from the Oxford English Dictionary (OED), e.g.\@ using `obsolete' tags. The shift start points were estimated using attestation dates in the OED.  The first six examples are words that shifted dramatically in meaning while the remaining four are words that acquired new meanings (while potentially also keeping their old ones). }
\label{testwords}
\end{table*}

\newcommand{\tbf}{\textbf}
First, we tested whether the methods capture known historical shifts in meaning.
The goal in this task is for the methods to correctly capture whether pairs of words moved closer or further apart in semantic space during a pre-determined time-period. 
We use a set of independently attested shifts as an evaluation set (Table \ref{testwords}). 
For comparison, we evaluated the methods on both the large (but messy) \engall data and the smaller (but clean) \coha data. 
On this task, all the methods performed almost perfectly in terms of capturing the correct directionality of the shifts (i.e., the pairwise similarity series have the correct sign on their Spearman correlation with time), but there were some differences in whether the methods deemed the shifts statistically significant at the $p<0.05$ level.\footnote{All subsequent significance tests are at $p<0.05$.} 
	Overall, SGNS performed the best on the full English data, but its performance dropped significantly on the smaller \coha dataset, where SVD performed best. 
	PPMI was noticeably worse than the other two approaches  (Table \ref{falseomission}).

\paragraph{Discovering shifts from data.}
\begin{table}
\centering
{\small
\begin{tabular}{llrr} 
\toprule  
Method & Corpus & \% Correct & \%Sig. \\ 
\midrule 
\multirow{2}{0.7cm}{PPMI} & \engall & 77.1 & 51.9\\
& \coha & 85.7 & 52.4\\
\multirow{2}{0.7cm}{SVD} & \engall & 92.6 & 81.5\\
& \coha & 95.8  & 62.5  \\
\multirow{2}{0.7cm}{SGNS} & \engall & 100.0 & 88.9\\
& \coha & 87.5 & 50.0\\
\bottomrule
\end{tabular}
}
\caption{Performance on detection task, i.e. ability to capture the attested shifts from Table \ref{testwords}. SGNS performs the best on the \engall\ corpus, whereas SVD performs the best on \textsc{coha}. \textbf{Note:} These results use an improved and corrected experimental protocol compared to earlier versions of this work. The general trends are consistent, but the absolute numbers for all methods are lower. See the Appendix for details, and please use these revised numbers for future comparisons.} 
\label{falseomission}
\end{table}
	We tested whether the methods discover reasonable shifts by examining the top-10 words that changed the most from the 1900s to the 1990s according to the semantic displacement metric introduced in Section \ref{displacement} (limiting our analysis to words with relative frequencies above $10^{-5}$ in both decades).
	We used the \engfic data as the most-changed list for \engall was dominated by scientific terms due to changes in the corpus sample.
	%; the COHA data was too small to robustly capture many shifts. 
		\begin{table*}
\centering
{\small
\begin{tabular}{lll} 
\toprule  
Method & Top-10 words that changed from 1900s to 1990s\\ 
\midrule
PPMI  & \underline{know}, \underline{got}, \underline{would}, \underline{decided}, \underline{think}, \underline{stop}, \underline{remember}, \tbf{started}, \underline{must}, \underline{wanted}\\
SVD &  harry, \tbf{headed}, \tbf{calls}, \tbf{gay}, 
wherever, \underline{male}, \tbf{actually}, special, cover, \underline{naturally}\\
SGNS & \tbf{wanting}, \tbf{gay}, \tbf{check}, \tbf{starting}, \tbf{major}, \tbf{actually}, \underline{touching}, harry, \tbf{headed}, romance\\
\bottomrule
\end{tabular} 
}
\caption{Top-10 English words with the highest semantic displacement values between the 1900s and 1990s.  Bolded entries correspond to real semantic shifts, as deemed by examining the literature and their nearest neighbors; for example, \textit{headed} shifted from primarily referring to the ``top of a body/entity'' to referring to ``a direction of travel.''
%\footnote{This shift, along with the shift for \textit{actually} described in Section \ref{tradling}, are both classic examples of subjectification, where words shift towards referring to the speakers internal mental states \cite{traugott_regularity_2001}.}.
	Underlined entries are borderline cases that are largely due to global genre/discourse shifts; for example, \textit{male} has not changed in meaning, but its usage in discussions of ``gender equality'' is relatively new.
	Finally, unmarked entries are clear corpus artifacts; for example, \textit{special}, \textit{cover}, and \textit{romance} are artifacts from the covers of fiction books occasionally including advertisements etc. }
\label{discovery}
\end{table*}
\begin{table*}
\centering
{\small
\begin{tabular}{llp{2.15in}p{2.3in}} 
\toprule  
Word & Language & Nearest-neighbors in 1900s & Nearest-neighbors in 1990s\\ 
\midrule
wanting & English & {lacking, deficient, lacked, lack, needed} & {wanted, something, wishing, anything, anybody} \\
\addlinespace[3pt]
asile & French & {refuge, asiles,  hospice, vieillards, infirmerie } & {demandeurs,  refuge, hospice,  visas,  admission}\\
\addlinespace[3pt]
widerstand & German & {scheiterte, volt, stromstärke, leisten, brechen} & {opposition, verfolgung, nationalsozialistische, nationalsozialismus, kollaboration}\\
\bottomrule
\end{tabular} 
}
\caption{Example words that changed dramatically in meaning in three languages, discovered using SGNS embeddings. The examples were selected from the top-10 most-changed lists between 1900s and 1990s as in Table \ref{discovery}. In English, \textit{wanting} underwent subjectification and shifted from meaning ``lacking'' to referring to subjective ''desire'', as in  ``the education system is wanting'' (1900s) vs. "I've been wanting to tell you'' (1990s). In French \textit{asile} (``asylum'') shifted from primarily referring to ``hospitals, or infirmaries'' to also referring to ``asylum seekers, or refugees''. Finally, in German \textit{Widerstand} (``resistance'') gained a formal meaning as referring to the local German resistance to Nazism during World War II.}
\label{examples}
\end{table*}
	Table \ref{discovery} shows the top-10 words discovered by each method. 
	These shifts were judged by the authors as being either clearly genuine, borderline, or clearly corpus artifacts. 
	SGNS performed by far the best on this task, with $70\%$ of its top-10 list corresponding to genuine semantic shifts, followed by $40\%$ for SVD, and $10\%$ for PPMI. 
	However, a large portion of the discovered words for PPMI (and less so SVD) correspond to borderline cases, e.g.\@ $\textit{know}$, that have not necessarily shifted significantly in meaning but that occur in different contexts due to global genre/discourse shifts. 
The poor quality of the nearest neighbors generated by the PPMI algorithm---which are skewed by PPMI's sensitivity to rare events---also made it difficult to assess the quality of its discovered shifts. 
SVD was the most sensitive to corpus artifacts (e.g., co-occurrences due to cover pages and advertisements), but it still captured a number of genuine semantic shifts. 

We opted for this small evaluation set and relied on detailed expert judgments to minimize ambiguity; each potential shift was analyzed in detail by consulting consulting existing literature (especially the OED; \citealt{simpson_oxford_1989}) and all disagreements were discussed. 

Table \ref{examples} details representative example shifts in English, French, and German.
Chinese lacks sufficient historical data for this task, as only years 1950-1999 are usable; however, we do still see some significant changes for Chinese in this short time-period, such as \begin{CJK*}{UTF8}{gbsn} 病毒 \end{CJK*}  (``virus'') moving closer to\begin{CJK*}{UTF8}{gbsn} 电脑 \end{CJK*} (``computer'', $\rho=0.89$).

%\subsubsection*{Smoothness}
%
%\begin{itemize}
%\item
%	Assuming that most words to not change drastically from decade to decade, we evaluate the smoothness of 1000 random pair-wise similarity plots.
%\item
%	Smoothness is quantified as the coefficient of variation of the first-difference of the similarity time-series, so series that change in a smooth fashion will have low values while series that tend to have large variable transitions will have high values.  
%\item
%	We find no significant difference between SVD and SGNS, despite the fact that SGNS is incrementally initialized. 
%\item
%	PPMI is significantly less smooth than the other two ($p<10^{-5}$) by pair-wise Mann-Whitney U-tests with Bonferroni corrections. 
%\end{itemize}

\subsection{Methodological recommendations}
PPMI is clearly worse than the other two methods; 
it performs poorly
on all the benchmark tasks, is extremely sensitive to rare events,
and is prone to false discoveries from global genre shifts.  
Between SVD and SGNS the results are somewhat equivocal, as both perform best on two out of the four tasks (synchronic accuracy, \engall detection, \coha detection, discovery).
Overall, SVD performs best on the synchronic accuracy task and has higher average accuracy on the `detection' task, while SGNS performs best on the `discovery' task. 
These results suggest that both these methods are reasonable choices for studies of semantic change but that they each have their own tradeoffs:
SVD is more sensitive, as it performs well on detection tasks even when using a small dataset, but this sensitivity also results in false discoveries due to corpus artifacts.
In contrast, SGNS is robust to corpus artifacts in the discovery task, but it is not sensitive enough to perform well on the detection task with a small dataset. 
Qualitatively, we found SGNS to be  most useful for discovering new shifts and visualizing changes (e.g., Figure \ref{wordpaths}), while SVD was most effective for detecting subtle shifts in usage.

\newcommand{\appropto}{\mathrel{\vcenter{
  \offinterlineskip\halign{\hfil$##$\cr
    \propto\cr\noalign{\kern2pt}\sim\cr\noalign{\kern-2pt}}}}}
\section{Statistical laws of semantic change}
\begin{table*}[t!]
\centering
{\small
\begin{tabular}{l|p{3.81in}} 
\toprule  
Top-10 most polysemous &  yet, always, even, little, called, also, sometimes, great, still, quite\\ 
\addlinespace[3pt]
Top-10 least polysemous & {photocopying, retrieval, thirties, mom, sweater, forties, seventeenth, fifteenth, holster, postage}\\
\bottomrule
\end{tabular} 
}
\caption{The top-10 most and least polysemous words in the \engfic data. Words like \textit{yet}, \textit{even}, and \textit{still} are used in many diverse ways and are highly polysemous. In contrast, words like \textit{photocopying}, \textit{postage}, and \textit{holster} tend to be used in very specific well-clustered contexts, corresponding to a single sense; for example, \textit{mail} and \textit{letter} are both very likely to occur in the context of \textit{postage} and are also likely to co-occur with each other, independent of \textit{postage}.}
\label{diversity}
\end{table*}
\begin{figure*}[t!]
\centering
\includegraphics[width=\textwidth]{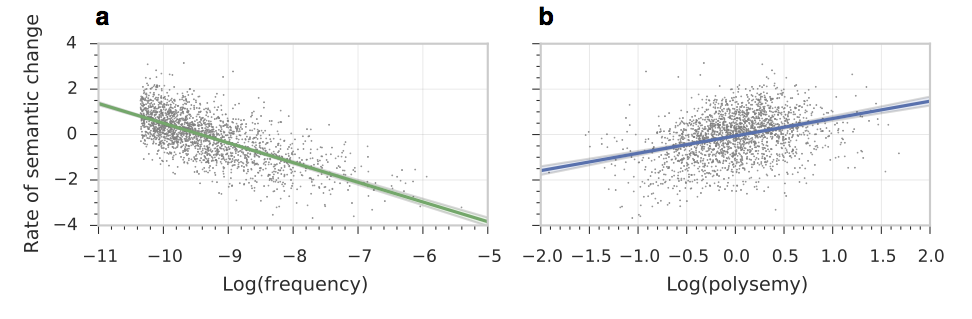}
\vspace{-15pt}
\caption{Higher frequency words have lower rates of change (\textbf{a}), while polysemous words have higher rates of change (\textbf{b}). The plots show robust linear regression fits \cite{huber_robust_2011} with 95\% CIs on the 2000s decade of the \coha\-lemma data.} 
\label{scatter}
\end{figure*}

We now show how diachronic embeddings can be used in a large-scale cross-linguistic analysis to reveal statistical laws that relate frequency and polysemy to semantic change. 
In particular, we analyze how a word's rate of semantic change, 
\begin{equation}
\Delta^{(t)}(w_i)=\textrm{cos-dist}(\mb{w}_i^{(t)},\mb{w}_i^{(t+1)})
\end{equation}
 depends on its frequency, $f^{(t)}(w_i)$ and a measure of its polysemy, $d^{(t)}(w_i) $ (defined in Section \ref{innovation}).% which quantifies the extent to which it appears in diverse contexts. 

\subsection{Setup}

We present results using SGNS embeddings.
Using all four languages and all four conditions for English (\textsc{EngAll}, \textsc{EngFic}, and \coha with and without lemmatization),
we performed regression analysis on rates of semantic change, $\Delta^{(t)}(w_i)$;
thus, we examined one data-point per word for each pair of consecutive decades and analyzed how a word's frequency and polysemy at time $t$ correlate with its degree of semantic displacement over the next decade.  
To ensure the robustness of our results, we analyzed only non-stop words that occurred more than 500 times in both decades contributing to a change (lower-frequency words tend to lack sufficient co-occurrence data across years). 
We also log-transformed the semantic displacement scores and normalized the scores to have zero mean and unit variance; we denote these normalized scores by $\tilde{\Delta}^{(t)}(w_i)$.

Though SGNS and SVD embeddings performed similarly in our evaluation tasks, we opted to use the SGNS embeddings since they provide a better estimate of the relationship between frequency and semantic change. 
With SVD embeddings the effect of frequency is confounded by the fact that high frequency words have less finite-sample variance in their co-occurrence estimates, which makes the word vectors of high frequency words appear more stable between corpora, regardless of any real semantic change. 
The SGNS embeddings do not suffer from this issue because they are initialized with the embeddings of the previous decade.\footnote{In fact, the SGNS embeddings may even be biased in the other direction, since higher frequency words undergo more SGD updates ``away'' from this initialization.} 

We performed our analysis using a linear mixed model with random intercepts per word and fixed effects per decade; i.e., we fit $\beta_{f}$,  $\beta_{d}$, and $\beta_t$ s.t.
\begin{multline}\label{mixed}
\tilde{\Delta}^{(t)}(w_i) = \beta_{f}\log\left(f^{(t)}(w_i)\right) + \beta_{d}\log\left(d^{(t)}(w_i)\right)\\ + \beta_{t} + z_{w_i} + \epsilon^{(t)}_{w_i} \:\:\forall w_i \in \mathcal{V},t\in \{t_0,...,t_n\},
\end{multline}
where $z_{w_i} \sim \mathcal{N}(0, \sigma_{w_i})$ is the random intercept for word $w_i$ and $\epsilon^{(t)}_{w_i} \in \mathcal{N}(0,\sigma)$ is an error term. 
$\beta_f,\beta_d$ and $\beta_t$ correspond to the fixed effects for frequency, polysemy and the decade $t$, respectively\footnote{Note that time is treated as a categorical variable, as each decade has its own fixed effect.}.
Intuitively, this model estimates the effects of frequency and polysemy on semantic change, while controlling for temporal trends and correcting for the fact that measurements on same word will be correlated across time. 
We fit \eqref{mixed} using the standard restricted maximum likelihood algorithm (\citealt{mcculloch_generalized_2001}; Appendix C).
%, which facilitates comparisons across datasets and methods (see Appendix C).

\subsection{Overview of results}

We find that, across languages, rates of semantic change obey a scaling relation of the form
\begin{equation}
\Delta(w_i) \propto f(w_i)^{\beta_{f}} \times d(w_i)^{\beta_{d}},
\end{equation}
with $\beta_{f}<0$ and $\beta_{d} > 0$.
This finding implies that frequent words change at slower rates while polysemous words change faster, and that both these relations scale as power laws.

\subsection{Law of conformity: Frequently used words change at slower rates}

Using the model in equation \eqref{mixed}, we found that the logarithm of a word's frequency, $\log(f(w_i))$, has a significant and substantial negative effect on rates of semantic change in all settings (Figures 2a  and 3a). 
Given the use of log-transforms in pre-processing the data this implies rates of semantic change are proportional to a negative power ($\beta_{f}$) of frequency, i.e.\@
\begin{equation}
\Delta(w_i) \propto f(w_i)^{\beta_f},
\end{equation}
 with $\beta_{f} \in [-1.24, -0.30]$ across languages/datasets. 
% Of course, some of this effect is due to the simple fact that the co-occurrence statistics for higher-frequency words contain less finite-sample variance; however, simulations show that this variance reduction effect cannot fully explain the observed correlations (Appendix D). 
 %The relatively large range of values for $\beta_{f}$ is due to the fact that the \coha datasets are outliers due to their substantially smaller sample sizes (Figure \ref{bar}; the range is $\beta_f \in [-0.66, -0.27]$ with \coha excluded). 
%Equivalently, since the rates of semantic change are z-scored, this means that for every order of magnitude increase in frequency, a word's rate of semantic change decreases by ${\sim}[0.3, 1.4]$ standard deviations. 

%This result compliments previous findings showing that rates of historical verb regularization \cite{lieberman_quantifying_2007} and lexical replacement \cite{pagel_frequency_2007} are also proportional to an inverse power of frequency.
%Together these results suggest a general tendency of linguistic conformity where frequently used words are less likely to change, a tendency which echoes the general notion of a conformity bias in cultural evolution \cite{boyd_culture_1988}. 

\subsection{Law of innovation: Polysemous words change at faster rates}\label{innovation}
%\footnotetext{The \coha data is ${\sim}100{\times}$ smaller, which has a global effect on the construction of the co-occurrence network (e.g., lower average degree) used to compute polysemy scores.}

There is a common hypothesis in the linguistic literature that ``words become semantically extended by being used in diverse contexts'' \cite{winter_cognitive_2014}, an idea that dates back to the writings of \newcite{breal_essai_1897}. 
We tested this notion by examining the relationship between polysemy and semantic change in our data. 

\subsubsection*{Quantifying polysemy}

\begin{figure*}
\centering
\vspace{-10pt}
\includegraphics[width=\textwidth]{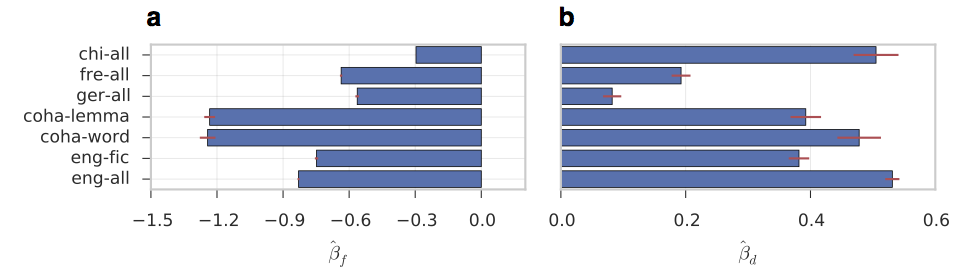}
\vspace{-10pt}
\caption{\textbf{a}, The estimated linear effect of $\log$-frequency ($\hat{\beta}_{f}$) is significantly negative across all languages. From the \coha data, we also see that the result holds regardless of whether lemmatization is used. \textbf{b}, Analogous trends hold for the linear effect of the polysemy score ($\hat{\beta}_{d}$), which is significantly positive across all conditions. The magnitudes of $\hat{\beta}_{f}$ and $\hat{\beta}_{d}$ vary significantly across languages, indicating language-specific variation within the general scaling trends.  95\% CIs are shown.}
\label{bar}
\end{figure*}

Measuring word polysemy is a difficult and fraught task, as even ``ground truth'' dictionaries  differ in the number of senses they assign to words \cite{simpson_oxford_1989,fellbaum_wordnet_1998}.
We circumvent this issue by measuring a word's {\em contextual diversity} as a proxy for its polysemousness. 
The intuition behind our measure is that words that occur in many distinct, unrelated contexts will tend to be highly polysemous. 
This view of polysemy also fits with previous work on semantic change, which emphasizes the role of contextual diversity \cite{breal_essai_1897,winter_cognitive_2014}. 

We measure a word's contextual diversity, and thus polysemy, by examining its neighborhood in an empirical co-occurrence network. 
We construct empirical co-occurrence networks for the top-10,000 non-stop words of each language using the PPMI measure defined in Section \ref{methods}.
In these networks words are connected to each other if they co-occur more than one would expect by chance (after smoothing). 
The polysemy of a word is then measured as its local clustering coefficient within this network \cite{watts_collective_1998}:
\begin{equation}
d(w_i) = -\frac{\sum_{c_i, c_j \in N_{\pmi}(w_i)}\ind\left\lbrace \pmi(c_i, c_j) > 0\right\rbrace}{|N_{\pmi}(w_i)|(|N_{\pmi}(w_i)|-1)},
\end{equation}
where $N_{\pmi}(w_i) = \lbrace w_j : \pmi(w_i, w_j) > 0\rbrace$.
This measure counts the proportion of $w_i$'s neighbors that are also neighbors of each other.
According to this measure, a word will have a high clustering coefficient (and thus a low polysemy score) if the words that it co-occurs with also tend to co-occur with each other.
Polysemous words that are contextually diverse will have low clustering coefficients, since they appear in disjointed or unrelated contexts.

Variants of this measure are often used in word-sense discrimination  and correlate
with, e.g., number of senses in WordNet \cite{dorow_discovering_2003,ferret_discovering_2004}.
However, we found that it was slightly biased towards rating contextually diverse discourse function words (e.g., \textit{also}) as highly polysemous, which needs to be taken into account when interpreting our results.
We opted to use this measure, despite this bias, because it has the strong benefit of being clearly interpretable: it simply measures the extent to which a word appears in diverse textual contexts. 
%Of course, this measure does not directly measure polysemy, as it does not distinguish between homonymy and genuine polysemy; however, in our investigations, we found very few cases where it was misled by a clearly homonymic relationship.
Table \ref{diversity} gives examples of the least and most polysemous words in the \engfic data, according to this score. 

As expected, this measure has significant intrinsic positive correlation with frequency.
Across datasets, we found Pearson correlations in the range $0.45 < r <  0.8$ (all $p<0.05$), confirming frequent words tend to be used in a greater diversity of contexts.
As a consequence of this high correlation,  we interpret the effect of this measure only after controlling for frequency (this control is naturally captured in equation \eqref{mixed}).

\subsubsection*{Polysemy and semantic change}

After fitting the model in equation \eqref{mixed}, we found that the logarithm of the polysemy score exhibits a strong positive effect on rates of semantic change, throughout all four languages (Figure 3b). 
As with frequency, the relation takes the form of a power law
\begin{equation}
\Delta(w_i) \propto d(w_i)^{\beta_{d}},
\end{equation}
with a language/corpus dependent scaling constant in $\beta_{d} \in [0.08, 0.53]$.
The distribution of polysemy scores varies substantially across languages, so the large range for this constant is not surprising.\footnote{For example, the \engall\ polysemy scores have an excess kurtosis that is $25\%$ larger than \textsc{GerAll}.}

Note that this relationship between polysemy and semantic change is a complete reversal from what one would expect according to $d(w_i)$'s positive correlation with frequency; i.e., since frequency and polysemy are highly positively correlated, one would expect them to have similar effects on semantic change, but we found that the effect of polysemy completely reversed after controlling for frequency. 
Figure 2b shows the relationship of polysemy with rates of semantic change in the \coha\-lemma data after regressing out effect of frequency (using the method of \citealt{graham_confronting_2003}).

\section{Discussion}

We show how distributional methods can
reveal statistical laws of semantic change and
offer a robust methodology for future work in this area.

Our work builds upon a wealth of previous research on quantitative approaches to semantic change, including prior work with distributional methods \cite{sagi_tracing_2011,wijaya_understanding_2011,gulordava_distributional_2011,jatowt_framework_2014,kulkarni_statistically_2014,xu_computational_2015}, as well as recent work on detecting the emergence of novel word senses \cite{lau_word_2012,mitra_thats_2014,cook_novel_2014,mitra_automatic_2015,frermann_bayesian_2016}.
We extend these lines of work by rigorously comparing different approaches to quantifying semantic change and by using these methods to propose new statistical laws of semantic change. 

The two statistical laws we propose have strong implications for future work in historical semantics. 
The {\em law of conformity}---frequent words change more slowly---clarifies frequency's role in semantic change.
Future studies of semantic change must account for frequency's conforming effect: when examining the interaction between some linguistic process and semantic change, the {\em law of conformity} should serve as a null model in which the interaction is driven primarily by underlying frequency effects. 

The {\em law of innovation}---polysemous words change more quickly---quantifies the central role polysemy plays in semantic change, an issue that has concerned linguists for more than 100 years \cite{breal_essai_1897}.
Previous works argued that semantic change leads to polysemy \cite{wilkins_part_1993,hopper_grammaticalization_2003}.
However, our results show that polysemous words change faster, which suggests that polysemy may actually lead to semantic change. 
%and formalizes an intuitive notion that has been hypothesized for decades: the more people use a word in different ways the more it will change. 

%Overall, these two factors---frequency and polysemy---explain between $48\%$ and $88\%$ of the variance\footnote{Marginal $R^2$ \cite{nakagawa_general_2013}.} in rates of semantic change (across conditions).
%This remarkable degree of explanatory power indicates that frequency and polysemy are perhaps the two most crucial linguistic factors that explain rates of semantic change over time. 

%These laws also inform future research 
%on the cognitive underpinnings of semantic change and 
%highlight parallels between sociolinguistic and biological evolution. 
These empirical statistical laws also lend themselves to various causal mechanisms.
The {\em law of conformity} might be a consequence of learning: perhaps
people are more likely to use rare words mistakenly in novel ways,
a mechanism formalizable by Bayesian models of word learning 
and corresponding to the biological notion of genetic drift \cite{reali_words_2010}. 
Or perhaps a sociocultural conformity bias makes people less likely to accept 
novel innovations of common words, a mechanism analogous to the biological 
process of purifying selection \cite{boyd_culture_1988,pagel_frequency_2007}. 
Moreover, such mechanisms may also be partially responsible for the {\em law of innovation}.
Highly polysemous words tend to have more rare senses \cite{kilgarriff},
and rare senses may be unstable by the {\em law of conformity}.
While our results cannot confirm such causal links, they nonetheless highlight a new role for
frequency and polysemy in language change and the importance of distributional
models in historical research.

\section*{Acknowledgments}
The authors thank D.\@ Friedman, R.\@ Sosic, C.\@ Manning, V. Prabhakaran, and S. Todd for their helpful comments and discussions.
We also thank S.\@ Tsutsui for catching a typo in equation \eqref{procrustes}, which is present in previous versions, and Astrid van Aggelen for catching transcription errors in previous versions of Tables 2 and 3. 
We are also indebted to our anonymous reviewers.
W.H. was supported by an NSERC PGS-D grant and the SAP Stanford Graduate Fellowship. W.H., D.J., and J.L. were supported by the Stanford Data Science Initiative, and NSF Awards IIS-1514268,  IIS-1149837, and IIS-1159679.
\bibliography{paper-hist_vec}
\bibliographystyle{acl2016}
\appendix
\section{Hyperparameter and pre-processing details}

For all datasets, words were lowercased and stripped of punctuation.
	For the Google datasets we built models using the top-100000 words by their average frequency over the entire historical time-periods, and we used the top-50000 for COHA. 
During model learning we also discarded all words within a year that occurred below a certain threshold (500 for the Google data, 100 for the COHA data).

For all methods, we used the hyperparameters recommended in \newcite{levy_improving_2015}. 
For the context word distributions in all methods, we used context distribution smoothing with a smoothing parameter of $0.75$.
Note that for SGNS this corresponds to smoothing the unigram negative sampling distribution. 
For both, SGNS and PPMI,  we set the negative sample prior $\alpha = \log(5)$, while we set this value to $\alpha=0$ for SVD, as this improved results. 
When using SGNS on the Google data, we also subsampled, with words being random removed with probability $p_r(w_i) = 1 - \sqrt{\frac{10^{-5}}{f(w_i)}}$, as recommended by \newcite{levy_improving_2015} and \newcite{mikolov_distributed_2013}.
Furthermore, to improve the computational efficiency of SGNS (which works with text streams and not co-occurrence counts), we downsampled the larger years in the Google N-Gram data to have at most $10^{9}$ tokens. 
No such subsampling was performed on the \coha data.

For all methods, we defined the context set to simply be the same vocabulary as the target words, as is standard in most word vector applications \cite{levy_improving_2015}.
However, we found that the PPMI method benefited substantially from larger contexts (similar results were found in \citealt{bullinaria_extracting_2007}), so we did not remove any low-frequency words per year from the context for that method. 
The other embedding approaches did not appear to benefit from the inclusion of these low-frequency terms, so they were dropped for computational efficiency. 

For SGNS, we used the implementation provided in \newcite{levy_improving_2015}.
The implementations for PPMI and SVD are released with the code package associated with this work. 
\section{Visualization algorithm}

To visualize semantic change for a word $w_i$ in two dimensions we employed the following procedure, which relies on the t-SNE embedding method \cite{van_der_maaten_visualizing_2008} as a subroutine:
\begin{enumerate}
\item
	Find the union of the word $w_i$'s $k$ nearest neighbors over all necessary time-points. 
\item
	Compute the t-SNE embedding of these words on the most recent (i.e., the modern) time-point. 
\item
	For each of the previous time-points, hold all embeddings fixed, except for the target word's (i.e., the embedding for $w_i$), and optimize a new t-SNE embedding only for the target word. We found that initializing the embedding for the target word to be the centroid of its $k'$-nearest neighbors in a time-point was highly effective. 
\end{enumerate}
Thus, in this procedure the background words are always shown in their ``modern'' positions, which makes sense given that these are the current meanings of these words. 
This approximation is necessary, since in reality all words are moving. 

\section{Regression analysis details}

In addition to the pre-processing mentioned in the main text, we also normalized the contextual diversity scores $d(w_i)$ within years by subtracting the yearly median. 
This was necessary because there was substantial changes in the median contextual diversity scores over years due to changes in corpus sample sizes etc.  
We removed stop words using the available lists in Python's NLTK package \cite{bird_natural_2009}. 
We follow \newcite{kim_temporal_2014} and allow a buffer period for the historical word vectors to initialize; we use a buffer period of four decades from the first usable decade and only measure changes after this period. 

When analyzing the effects of frequency and contextual diversity, the model contained fixed effects for these features and for time along with random effects for word identity. 
We opted not to control for POS tags in the presented results, as contextual diversity is co-linear with these tags (e.g., adverbs are more contextual diverse than nouns), and the goal was to demonstrate the main effect of contextual diversity across all word types.  
%That said, the effect of contextual diversity remained strong and significantly positive in all datasets even after controlling for POS tags. 

To fit the linear mixed models, we used the Python $\texttt{statsmodels}$ package with restricted maximum likelihood estimation (REML) \cite{statsmodels2010}. 
All mentioned significance scores were computed according to Wald's $z$-tests.
%, though these results agreed with Bonferroni corrected likelihood ratio tests on the $\texttt{eng-all}$ data. 

%The visualizations in Figure \ref{scatter} were computed on the \texttt{eng-all} data and correspond to bootstrapped locally-linear kernel regressions with bandwidths selected via the AIC Hurvitch criteria \cite{li_nonparametric_2007}.

\section{Revisions to the methodology for ``detecting known shifts'' (Table 3)}

The methodology for ``detecting known shifts'' has been improved to correct for certain issues, most prominently: 
\begin{itemize}
	\item 
		In earlier versions, the inclusion/exclusion of word pairs in particular time points based on frequency cutoffs was unnecessarily strict and not properly detailed. In the previous versions, we used the same cutoffs as for the analysis in Section 4 (i.e., frequencies had to be above $10^{-5}$), but this was not clear in the text. In this revised version, we compute cosine similarities for pairs of words in a time period if both are above the minimum count for the embedding construction (100 occurrences for \coha and 500 for the \engall corpus).
		This results in lower scores overall but is more reflective of how downstream users make use of our embeddings and reflects the exact results one obtains by running our off-the-shelf embeddings (available on the project website) through the evaluation. 
		For time points where one of the words in a pair is below the threshold, we simply discard these time points from the Spearman correlation. 
	\item 
		In earlier versions, Spearman correlations were computed for pairs with less than 5 time points. However, we now require at least 5 time points to have a minimum amount of robustness.  
	\item 
		In earlier versions, the SGNS model used the incorrect date for the start of the shift for the word \textit{gay}. 
\end{itemize}

A script for replicating the numbers in Table 3, using this revised methodology, is now available in the Github repo associated with this work. 
Note also that not all pairs in Table 2 are actually used for evaluation in all settings (e.g., for \textsc{coha}, due to not having enough samples).

\end{document}